\documentclass[runningheads]{llncs}
\usepackage[table,xcdraw]{xcolor}
\usepackage{graphicx}
\usepackage{adjustbox}
\usepackage{multirow}
\usepackage{sparklines}
\usepackage{booktabs}
\usepackage{sparklines}

\usepackage[labelfont=bf]{caption}

\usepackage{multirow}
\usepackage{array}
\usepackage{arydshln}
\def\sparkrectangleh #1 #2 {%
   \ifdim #1pt > #2pt
        \errmessage{The left corner #1 of rectangle cannot be lower than #2}%
   \fi
   {\pgfmoveto{\pgforigin}\color{sparkrectanglecolor}%
   \pgfrect[fill]{\pgfxy(#1, 0)}{\pgfxy(#2-#1,1)}}}%

\usepackage{array}{}
\usepackage[round-mode=places,detect-weight=true,detect-inline-weight=math]{siunitx}

\def\sparklineheight ex{7pt}
\usepackage{multicol}
\usepackage{amsmath}
\usepackage{amsfonts}
\usepackage{chngpage}
\usepackage{todonotes} 
\usepackage[multiple]{footmisc}

\usepackage[hyphens]{url}
\usepackage{hyperref}
\usepackage[hyphenbreaks]{breakurl}
\usepackage{breakcites}

\begin{document}
%
\title{Quantifying Valence and Arousal in Text \newline with Multilingual Pre-trained \newline Transformers }

%

\titlerunning{Quantifying Valence and Arousal in Text with Multilingual Transformers}

\author{Gonçalo Azevedo Mendes\inst{1,2}\orcidID{0000-0002-0595-3367} \and Bruno Martins\inst{1,2}\orcidID{0000-0002-3856-2936}} 
\authorrunning{G. A. Mendes and B. Martins}

\institute{Instituto Superior Técnico, Universidade de Lisboa, Lisboa, Portugal \and INESC-ID, Lisboa, Portugal \email{\{goncalo.a.mendes,bruno.g.martins\}@tecnico.ulisboa.pt}}

\maketitle              
\begin{abstract}\label{section:abstract}
The analysis of emotions expressed in text has numerous applications. In contrast to categorical analysis, focused on classifying emotions according to a pre-defined set of common classes, dimensional approaches can offer a more nuanced way to distinguish between different emotions. Still, dimensional methods have been less studied in the literature. Considering a valence-arousal dimensional space, this work assesses the use of pre-trained Transformers to predict these two dimensions on a continuous scale, with input texts from multiple languages and domains. We specifically combined multiple annotated datasets from previous studies, corresponding to either emotional lexica or short text documents, and evaluated models of multiple sizes and trained under different settings. Our results show that model size can have a significant impact on the quality of predictions, and that by fine-tuning a large model we can confidently predict valence and arousal in multiple languages. We make available the code, models, and supporting data.

\keywords{Transformer-Based Multilingual Language Models \and Emotion Analysis in Text \and Predicting Valence and Arousal}
\end{abstract}

\section{Introduction}   \label{section:introduction}

The task of analyzing emotions expressed in text is commonly modeled as a classification problem, representing affective states (e.g., Ekman's six basic emotions~\cite{ekman1992}) as specific classes. 
The alternative approach of dimensional emotion analysis focuses on rating emotions according to a pre-defined set of dimensions, offering a more nuanced way to distinguish between different emotions \cite{Buechel_Emotion_Analysis_regProb}. Emotional states are represented on a continuous numerical space, with the most common dimensions defined as valence and arousal. In particular, valence describes the pleasantness of a stimulus, ranging from negative to positive feelings. Arousal represents the degree of excitement provoked by a stimulus, from calm to excited. The Valence-Arousal (VA) space~\cite{sam} corresponds to a 2-dimensional space to which a text sequence can be mapped. 

This study proposes using pre-trained multilingual Transformer models to predict valence and arousal ratings in text from different languages and domains. Models pre-trained on huge amounts of data from multiple languages can be fine-tuned to different types of downstream tasks with relatively small datasets in one or few languages, and still obtain reliable results on different languages~\cite{pires-etal-2019-multilingual}.
While previous research focused on monolingual VA prediction as regression from text, this study compiled 34 publicly available psycho-linguistic datasets, from different languages, into a single uniform dataset. 
We then eva luated multilingual DistilBERT~\cite{Sanh2019DistilBERTAD} and XLM-RoBERTa \cite{xlmr-conneau-etal-2020-unsupervised} models, to understand the impact of model size and training conditions on the ability to correctly predict affective ratings from textual contents.

Experimental results show that multilingual VA prediction is possible with a single Transformer model, particularly when considering the larger XLM-RoBERTa model. Even if performance differs across languages, most results improve or stay in line with the results from previous research focused on predicting these affective ratings on a single language. The code, models, and data used in this study are available on a GitHub repository\footnote{
\url{https://www.github.com/gmendes9/multilingual\_va\_prediction}
}.

The rest of the paper is organized as follows: Section \ref{section:related_work} presents related work, while Section \ref{section:models} describes the models considered for predicting valence and arousal. Section \ref{section:resources} describes the corpora used for model training and evaluation. Section \ref{section:exp_results} presents our findings and compares the results. Finally, Section \ref{section:conclusions} summarizes the main findings and discusses possibilities for future work.


\section{Related Work}\label{section:related_work}

Since Russel~\cite{Russel1980} first proposed a two-dimensional model of emotions, based on valence and arousal, much research has been done on dimensional emotion analysis. Most relevant to this study are the main lexicons~\cite{anew,dataset_nrc-vad,dataset_angst,dataset_spanish_norms,dataset_english_norms} and corpora~\cite{dataset_anet,dataset_emobank1,dataset_mas,dataset_cvaw_cvat} annotated according to these dimensions, used in previous work. Still, while several NLP and IR studies have addressed dimensional emotion extraction, most previous work has focused on categorical approaches~\cite{alm-etal-2005-emotions}.

Trying to predict valence and arousal has long been a relevant topic, both at the word-level~\cite{emotion_lexicons_91_lang,Du_pred,Hollis,Recchia,sedoc-etal-2017-predicting,wu-etal-2017,Yu_Pipelined_NN} and at the sentence/text-level~\cite{Buechel_Emotion_Analysis_regProb,Buechel_A_flexible_map,Kratzwald2018pred,dataset_chinese_emobank,Paltoglou,dataset_facebook_posts,Akhtar_PREDICTION,Wang2020,dataset_cvai}. Recchia et al. used pointwise mutual information coupled with k-NN regression to estimate valence and arousal for words~\cite{Recchia}. Hollis et al. resorted to linear regression modelling~\cite{Hollis}. Sedoc et al. combined distributional approaches with signed spectral clustering~\cite{sedoc-etal-2017-predicting}. Du and Zhang explored the use of CNNs~\cite{Du_pred}. Wu et al. used a densely
connected LSTM network and word features to identify emotions on the VA space for words and phrases~\cite{wu-etal-2017}. 
More recently, Buechel et al. proposed a method for creating arbitrarily large emotion lexicons in 91 languages, using a translation model, a target language embedding model, and a multitask learning feed-forward neural network~\cite{emotion_lexicons_91_lang}. This last work is interesting when compared to ours, as it is one of the few attempts to predict VA at a multilingual level, if only for individual words.

Paltoglou et al. attempted text-level VA prediction by resorting to affective dictionaries, as supervised machine learning techniques were inadequate for the small dataset used in their tests~\cite{Paltoglou}. Preo{\c{t}}iuc-Pietro et al. compiled a corpus of Facebook posts and built a bag-of-words (BoW) linear regression prediction model~\cite{dataset_facebook_posts}. Similarly, Buechel and Hahn used BoW representations in conjunction with TF-IDF weights~\cite{Buechel_Emotion_Analysis_regProb,Buechel_A_flexible_map}. More recently, several studies have compared CNNs and RNNs, amongst other neural architectures~\cite{Kratzwald2018pred,Akhtar_PREDICTION,Wang2020,dataset_cvai}. For instance, Lee et al. explored different methods for prediction, ranging from linear regression to multiple neural network architectures~\cite{dataset_chinese_emobank}. This last study explored the use of a BERT model, but differs from our work as the data is not multilingual.
The present work follows the steps of some of the aforementioned studies leveraging deep learning, aiming to build a single multilingual model capable of predicting affective ratings for valence and~arousal.

\section{Models for Predicting Valence and Arousal from Text}\label{section:models}

We address the prediction of valence and arousal scores as text-based regression, using pre-trained multilingual models adapted from the Huggingface library \cite{wolf-2020-transformers-huggingface}. In particular, we use DistilBERT \cite{Sanh2019DistilBERTAD} and XLM-RoBERTa \cite{xlmr-conneau-etal-2020-unsupervised} models. 

The multilingual DistilBERT model, consisting of 134M parameters, is based on a 6 layer Transformer encoder, with 12 attention heads and a hidden state size of 768. The model can train two times faster with only a slight performance decrease (approx. 5\%), compared to a multilingual BERT-base model with 25\% more parameters. As for XLM-RoBERTa, we used both the base (270M parameters) and large (550M parameters) versions. The base version is a 12 layer Transformer, with 12 attention heads and a hidden state size of 768. The large version uses 24 layers, 16 attention heads, and a hidden state size of 1024.

Both these models are pre-trained on circa 100 different languages, which will likely enable the generalization to languages for which there are no annotated data in terms of valence and arousal ratings. These models are fine-tuned for the task at hand with a regression head on top, consisting of a linear layer on top of the pooled representation from the Transformer (i.e., the representation of the first token in the input~sequence). 

The regression head produces two outputs, which are processed through a hard sigmoid activation function, forcing the predicted values on both dimensions to respect the target interval between zero and one.

Three loss functions were initially compared for model training, namely the Mean Squared Error (MSE), the Concordance Correlation Coefficient Loss (CCCL), and a recently proposed Robust Loss (RL) function~\cite{robust-loss}. In all these cases, the models are trained with the sum of the loss for the valence and arousal predictions, equally weighting both affective dimensions. 

MSE is the most used loss function in regression problems and can be defined as the mean of the squared differences between predicted ($\hat{y}$) and ground-truth~($y$) values, as shown in Equation \ref{eqn: loss_mse}. 

\begin{equation}
    \mathrm{MSE} = \frac{1}{N}\sum_{i=0}^{N}(y_i-\hat{y_i})^{2}.
    \label{eqn: loss_mse}
\end{equation}

The CCCL corresponds to a correlation-based function, evaluating the ranking agreement between the true and predicted values, within a batch of instances. It varies from the Pearson correlation by penalizing the score in proportion to the deviation if the predictions shift in value. Atmaja and Akagi \cite{ccc-loss} compared this function to the MSE and Mean Absolute Error (MAE) loss functions for the task of predicting emotional ratings from speech signals using LSTM neural networks, suggesting that this loss yields a better performance than error-based functions. The CCCL follows Equation \ref{eqn: loss_ccc_1}, where $\rho_{y\hat{y}}$ represents the Pearson correlation coefficient between $y$ and $\hat{y}$, $\sigma$ represents the standard deviation, and $\mu$ the mean value. Notice that the correlation ranges from -1 to 1, and thus we use one minus the correlation as the loss. 

\begin{equation}
    \mathrm{CCCL} = 1-\frac{2\rho_{y\hat{y}}\sigma_y\sigma_{\hat{y}}}{\sigma_{y^{2}}+\sigma_{\hat{y}^{2}}+(\mu_{y}-\mu_{\hat{y}})^{2}}.
    \label{eqn: loss_ccc_1}
\end{equation}

The RL function generalizes some of the most common robust loss functions (e.g., the Huber loss), that reduce the influence of outliers~\cite{robust-loss}, being described in its general form through Equation \ref{eqn: loss_robust_gen}. In this function, $x$ is the variable being minimized, corresponding to the difference between true and predicted values (i.e., $x_i=y_i-\hat{y_i}$). The function involves two parameters that tune its shape, namely $\alpha \in \mathbb{R}$ that controls the robustness, and a scale parameter $c>0$ which controls the size of its quadratic~bowl. 
\begin{equation}
    \mathrm{RL}= \frac{1}{N}\sum_{i=0}^{N}
    \begin{cases}
        \frac{1}{2}(x_i/c)^{2} & \text{ if } \alpha=2 \\ 
        \textup{log}\left ( \frac{1}{2}(x_i/c)^{2}+1 \right ) & \text{ if }\alpha=0 \\ 
        1-\textup{exp}\left ( -\frac{1}{2}(x_i/c)^{2} \right ) & \text{ if } \alpha=\infty \\ 
        \frac{|\alpha-2|}{\alpha}\left ( \left (\frac{(x_i/c)^{2}}{|\alpha-2|}+1\right )^{\alpha/2}-1 \right ) & \text{otherwise}.
    \end{cases}
\label{eqn: loss_robust_gen}
\end{equation}

A lower value of $\alpha$ implies penalizing minor errors at the expense of larger ones, while a higher value of $\alpha$ allows more inliers while increasing the penalty for outliers. 
We used the adaptive form of this robust loss function, where the parameter $\alpha$ is optimized and tuned during model training via stochastic gradient descent, as explained in the original 
paper~\cite{robust-loss}.

We also tested two hybrid loss functions derived from the previous ones, combining their different properties and merits. 
While the MSE and the RL functions analyze results at the instance level, the CCCL function does the same at the batch level. With this in mind, one hybrid loss function combines the CCCL and the MSE functions, while the other combines the CCCL with the RL function, in both cases through a simple addition.

\section{Resources}\label{section:resources}

We collected 34 different public datasets to form a large corpus of annotated data for the emotional dimensions of valence and arousal, with the intent to build the largest possible multilingual dataset. The original datasets comprise 13 different languages, which represent up to 2.5 billion native speakers worldwide\footnote{\url{https://www.cia.gov/the-world-factbook/countries/world/\#people-and-society}}\footnote{\url{https://www.ethnologue.com/}}. 
There are two types of datasets described on Table \ref{tab:ds_characterization}, namely word and short text datasets, respectively associating valence and arousal ratings to either individual words or short text sequences. All of these datasets were manually annotated by humans, either via crowdsourcing or by experienced linguists/psychologists, according to the Self-Assessment Manikin~(SAM) method~\cite{sam}. In addition, several lexicons relate to the Affective Norms for English Words (ANEW) resource, corresponding to either adaptations to other languages or extensions in terms of the number of words~\cite{anew}. ANEW was the first lexicon providing real-valued scores for the emotional dimensions of valence and arousal. It is important to note that this lexicon is excluded from our corpus for being part of larger datasets that were included, such as the one from Warriner et al.~\cite{dataset_english_norms}.  

\begin{table*}[!tb]
\centering
\caption{Dataset characterization. $\mu_{\tiny{\textrm{length}}}$ represents the mean text length of each instance, in terms of the number of words. $\mu$ and $\sigma$ represent the mean and standard deviation, in the emotional ratings, respectively.}
\label{tab:ds_characterization}
\scriptsize
\setlength{\tabcolsep}{2.9pt}
\begin{tabular}{llrrcccccc} 
\toprule
  & & & & \multicolumn{3}{c}{\footnotesize{\textbf{Arousal}}} & \multicolumn{3}{c}{\footnotesize{\textbf{Valence}}} \\ \cmidrule(lr){5-7} \cmidrule(lr){8-10}  
\footnotesize{\textbf{Dataset}}  & \footnotesize{Language} & \footnotesize{Items} & \footnotesize{$\mu_{\tiny{\textrm{length}}}$} & \footnotesize{$\mu{}$} & \footnotesize{$\sigma{}$} &  & \footnotesize{$\mu{}$} & \footnotesize{$\sigma{}$} &  \\ \midrule
\scriptsize{EmoBank}~\cite{dataset_emobank1,dataset_emobank2} & \scriptsize{English} & \scriptsize{10062} & \scriptsize{23.27} & \scriptsize{0.51} & \scriptsize{0.06} & 
\begin{sparkline}{10} \definecolor{sparkbottomlinecolor}{gray}{0.8} \setlength\sparkbottomlinethickness{0.7pt} \sparkbottomlinex -0.05 1.05 \sparkspike 0.0 0.0 \sparkspike 0.1111 0.0 \sparkspike 0.2222 0.0016 \sparkspike 0.3333 0.0395 \sparkspike 0.4444 0.5866 \sparkspike 0.5556 1.0 \sparkspike 0.6667 0.1083 \sparkspike 0.7778 0.0129 \sparkspike 0.8889 0.0026 \sparkspike 1.0 0.0 \end{sparkline}
& \scriptsize{0.49} & \scriptsize{0.09} & 
\begin{sparkline}{10} \definecolor{sparkbottomlinecolor}{gray}{0.8} \setlength\sparkbottomlinethickness{0.7pt} \sparkbottomlinex -0.05 1.05 \sparkspike 0.0 0.0005 \sparkspike 0.1111 0.004 \sparkspike 0.2222 0.0525 \sparkspike 0.3333 0.1507 \sparkspike 0.4444 0.4579 \sparkspike 0.5556 1.0 \sparkspike 0.6667 0.1177 \sparkspike 0.7778 0.021 \sparkspike 0.8889 0.0032 \sparkspike 1.0 0.0 \end{sparkline}
\\
\scriptsize{IEMOCAP}~\cite{dataset_iemocap} & \scriptsize{English} & \scriptsize{10039} & \scriptsize{19.22} & \scriptsize{0.56} & \scriptsize{0.22} & 
\begin{sparkline}{10} \definecolor{sparkbottomlinecolor}{gray}{0.8} \setlength\sparkbottomlinethickness{0.7pt} \sparkbottomlinex -0.05 1.05 \sparkspike 0.0 0.017 \sparkspike 0.1111 0.1827 \sparkspike 0.2222 0.6293 \sparkspike 0.3333 0.3892 \sparkspike 0.4444 0.0 \sparkspike 0.5556 0.829 \sparkspike 0.6667 1.0 \sparkspike 0.7778 0.9717 \sparkspike 0.8889 0.4501 \sparkspike 1.0 0.0312 \end{sparkline}
& \scriptsize{0.48} & \scriptsize{0.17} & 
\begin{sparkline}{10} \definecolor{sparkbottomlinecolor}{gray}{0.8} \setlength\sparkbottomlinethickness{0.7pt} \sparkbottomlinex -0.05 1.05 \sparkspike 0.0 0.0 \sparkspike 0.1111 0.0572 \sparkspike 0.2222 0.0739 \sparkspike 0.3333 0.3536 \sparkspike 0.4444 0.19 \sparkspike 0.5556 1.0 \sparkspike 0.6667 0.5849 \sparkspike 0.7778 0.1566 \sparkspike 0.8889 0.1073 \sparkspike 1.0 0.0044 \end{sparkline}
\\
\scriptsize{Facebook Posts}~\cite{dataset_facebook_posts} & \scriptsize{English} & \scriptsize{2894} & \scriptsize{28.15} & \scriptsize{0.29} & \scriptsize{0.25} & 
\begin{sparkline}{10} \definecolor{sparkbottomlinecolor}{gray}{0.8} \setlength\sparkbottomlinethickness{0.7pt} \sparkbottomlinex -0.05 1.05 \sparkspike 0.0 1.0 \sparkspike 0.1111 0.8179 \sparkspike 0.2222 0.2243 \sparkspike 0.3333 0.2968 \sparkspike 0.4444 0.1689 \sparkspike 0.5556 0.4459 \sparkspike 0.6667 0.2361 \sparkspike 0.7778 0.0369 \sparkspike 0.8889 0.0501 \sparkspike 1.0 0.0 \end{sparkline}
& \scriptsize{0.53} & \scriptsize{0.15} & 
\begin{sparkline}{10} \definecolor{sparkbottomlinecolor}{gray}{0.8} \setlength\sparkbottomlinethickness{0.7pt} \sparkbottomlinex -0.05 1.05 \sparkspike 0.0 0.0 \sparkspike 0.1111 0.0572 \sparkspike 0.2222 0.0739 \sparkspike 0.3333 0.3536 \sparkspike 0.4444 0.19 \sparkspike 0.5556 1.0 \sparkspike 0.6667 0.5849 \sparkspike 0.7778 0.1566 \sparkspike 0.8889 0.1073 \sparkspike 1.0 0.0044 \end{sparkline}
\\
\scriptsize{EmoTales}~\cite{dataset_emotales} & \scriptsize{English} & \scriptsize{1395} & \scriptsize{17.91} & \scriptsize{0.55} & \scriptsize{0.12} & 
\begin{sparkline}{10} \definecolor{sparkbottomlinecolor}{gray}{0.8} \setlength\sparkbottomlinethickness{0.7pt} \sparkbottomlinex -0.05 1.05 \sparkspike 0.0 0.0069 \sparkspike 0.1111 0.0 \sparkspike 0.2222 0.0361 \sparkspike 0.3333 0.1015 \sparkspike 0.4444 0.4596 \sparkspike 0.5556 1.0 \sparkspike 0.6667 0.5009 \sparkspike 0.7778 0.2048 \sparkspike 0.8889 0.0516 \sparkspike 1.0 0.0052 \end{sparkline}
& \scriptsize{0.49} & \scriptsize{0.15} & 
\begin{sparkline}{10} \definecolor{sparkbottomlinecolor}{gray}{0.8} \setlength\sparkbottomlinethickness{0.7pt} \sparkbottomlinex -0.05 1.05 \sparkspike 0.0 0.0057 \sparkspike 0.1111 0.0684 \sparkspike 0.2222 0.1977 \sparkspike 0.3333 0.2966 \sparkspike 0.4444 0.4791 \sparkspike 0.5556 1.0 \sparkspike 0.6667 0.3308 \sparkspike 0.7778 0.116 \sparkspike 0.8889 0.0437 \sparkspike 1.0 0.0 \end{sparkline}
\\
\scriptsize{ANET}~\cite{dataset_anet} & \scriptsize{English} & \scriptsize{120} & \scriptsize{31.96} & \scriptsize{0.66} & \scriptsize{0.22} & 
\begin{sparkline}{10} \definecolor{sparkbottomlinecolor}{gray}{0.8} \setlength\sparkbottomlinethickness{0.7pt} \sparkbottomlinex -0.05 1.05 \sparkspike 0.0 0.0 \sparkspike 0.1111 0.1471 \sparkspike 0.2222 0.2059 \sparkspike 0.3333 0.2353 \sparkspike 0.4444 0.1765 \sparkspike 0.5556 0.3529 \sparkspike 0.6667 0.3824 \sparkspike 0.7778 0.8824 \sparkspike 0.8889 1.0 \sparkspike 1.0 0.1471 \end{sparkline}
& \scriptsize{0.52} & \scriptsize{0.33} & 
\begin{sparkline}{10} \definecolor{sparkbottomlinecolor}{gray}{0.8} \setlength\sparkbottomlinethickness{0.7pt} \sparkbottomlinex -0.05 1.05 \sparkspike 0.0 0.32 \sparkspike 0.1111 1.0 \sparkspike 0.2222 0.32 \sparkspike 0.3333 0.0 \sparkspike 0.4444 0.04 \sparkspike 0.5556 0.28 \sparkspike 0.6667 0.36 \sparkspike 0.7778 0.08 \sparkspike 0.8889 0.84 \sparkspike 1.0 0.76   \end{sparkline} 
\\
\scriptsize{PANIG}~\cite{dataset_panig} & \scriptsize{German} & \scriptsize{619} & \scriptsize{9.12} & \scriptsize{0.47} & \scriptsize{0.12} & 
\begin{sparkline}{10} \definecolor{sparkbottomlinecolor}{gray}{0.8} \setlength\sparkbottomlinethickness{0.7pt} \sparkbottomlinex -0.05 1.05 \sparkspike 0.0 0.0 \sparkspike 0.1111 0.0 \sparkspike 0.2222 0.1881 \sparkspike 0.3333 0.7525 \sparkspike 0.4444 1.0 \sparkspike 0.5556 0.6337 \sparkspike 0.6667 0.3465 \sparkspike 0.7778 0.1337 \sparkspike 0.8889 0.0099 \sparkspike 1.0 0.0 \end{sparkline}
& \scriptsize{0.40} & \scriptsize{0.22} & 
\begin{sparkline}{10} \definecolor{sparkbottomlinecolor}{gray}{0.8} \setlength\sparkbottomlinethickness{0.7pt} \sparkbottomlinex -0.05 1.05 \sparkspike 0.0 0.0345 \sparkspike 0.1111 0.4885 \sparkspike 0.2222 1.0 \sparkspike 0.3333 0.523 \sparkspike 0.4444 0.2356 \sparkspike 0.5556 0.2414 \sparkspike 0.6667 0.2931 \sparkspike 0.7778 0.3276 \sparkspike 0.8889 0.1264 \sparkspike 1.0 0.0 \end{sparkline}
\\
\scriptsize{COMETA sentences}~\cite{dataset_cometa} & \scriptsize{German} & \scriptsize{120} & \scriptsize{16.75} & \scriptsize{0.48} & \scriptsize{0.15} & 
\begin{sparkline}{10} \definecolor{sparkbottomlinecolor}{gray}{0.8} \setlength\sparkbottomlinethickness{0.7pt} \sparkbottomlinex -0.05 1.05 \sparkspike 0.0 0.0 \sparkspike 0.1111 0.0882 \sparkspike 0.2222 0.4706 \sparkspike 0.3333 0.5588 \sparkspike 0.4444 0.6176 \sparkspike 0.5556 1.0 \sparkspike 0.6667 0.6176 \sparkspike 0.7778 0.1765 \sparkspike 0.8889 0.0 \sparkspike 1.0 0.0 \end{sparkline}
& \scriptsize{0.50} & \scriptsize{0.20} & 
\begin{sparkline}{10} \definecolor{sparkbottomlinecolor}{gray}{0.8} \setlength\sparkbottomlinethickness{0.7pt} \sparkbottomlinex -0.05 1.05 \sparkspike 0.0 0.0312 \sparkspike 0.1111 0.1562 \sparkspike 0.2222 0.375 \sparkspike 0.3333 1.0 \sparkspike 0.4444 0.375 \sparkspike 0.5556 0.4375 \sparkspike 0.6667 0.5938 \sparkspike 0.7778 0.5938 \sparkspike 0.8889 0.1875 \sparkspike 1.0 0.0 \end{sparkline}
\\
\scriptsize{COMETA stories}~\cite{dataset_cometa} & \scriptsize{German} & \scriptsize{64} & \scriptsize{90.17} & \scriptsize{0.53} & \scriptsize{0.15} & 
\begin{sparkline}{10} \definecolor{sparkbottomlinecolor}{gray}{0.8} \setlength\sparkbottomlinethickness{0.7pt} \sparkbottomlinex -0.05 1.05 \sparkspike 0.0 0.0 \sparkspike 0.1111 0.0 \sparkspike 0.2222 0.3333 \sparkspike 0.3333 0.5333 \sparkspike 0.4444 1.0 \sparkspike 0.5556 1.0 \sparkspike 0.6667 0.8667 \sparkspike 0.7778 0.3333 \sparkspike 0.8889 0.2 \sparkspike 1.0 0.0 \end{sparkline}
& \scriptsize{0.56} & \scriptsize{0.21} & 
\begin{sparkline}{10} \definecolor{sparkbottomlinecolor}{gray}{0.8} \setlength\sparkbottomlinethickness{0.7pt} \sparkbottomlinex -0.05 1.05 \sparkspike 0.0 0.0 \sparkspike 0.1111 0.0 \sparkspike 0.2222 0.3571 \sparkspike 0.3333 0.6429 \sparkspike 0.4444 0.4286 \sparkspike 0.5556 0.6429 \sparkspike 0.6667 0.2857 \sparkspike 0.7778 1.0 \sparkspike 0.8889 0.4286 \sparkspike 1.0 0.0714 \end{sparkline}
\\
\scriptsize{CVAT}~\cite{dataset_cvaw_cvat} & \scriptsize{Mandarin} & \scriptsize{2969} & \scriptsize{58.00} & \scriptsize{0.48} & \scriptsize{0.13} & 
\begin{sparkline}{10} \definecolor{sparkbottomlinecolor}{gray}{0.8} \setlength\sparkbottomlinethickness{0.7pt} \sparkbottomlinex -0.05 1.05 \sparkspike 0.0 0.0069 \sparkspike 0.1111 0.0446 \sparkspike 0.2222 0.2491 \sparkspike 0.3333 0.6286 \sparkspike 0.4444 0.8537 \sparkspike 0.5556 1.0 \sparkspike 0.6667 0.4686 \sparkspike 0.7778 0.1211 \sparkspike 0.8889 0.0206 \sparkspike 1.0 0.0 \end{sparkline}
& \scriptsize{0.48} & \scriptsize{0.17} & 
\begin{sparkline}{10} \definecolor{sparkbottomlinecolor}{gray}{0.8} \setlength\sparkbottomlinethickness{0.7pt} \sparkbottomlinex -0.05 1.05 \sparkspike 0.0 0.01 \sparkspike 0.1111 0.1605 \sparkspike 0.2222 0.7742 \sparkspike 0.3333 0.9582 \sparkspike 0.4444 0.6304 \sparkspike 0.5556 1.0 \sparkspike 0.6667 0.9666 \sparkspike 0.7778 0.4348 \sparkspike 0.8889 0.0301 \sparkspike 1.0 0.0 \end{sparkline}
\\
\scriptsize{CVAI}~\cite{dataset_cvai} & \scriptsize{Mandarin} & \scriptsize{1465} & \scriptsize{29.53} & \scriptsize{0.51} & \scriptsize{0.12} & 
\begin{sparkline}{10} \definecolor{sparkbottomlinecolor}{gray}{0.8} \setlength\sparkbottomlinethickness{0.7pt} \sparkbottomlinex -0.05 1.05 \sparkspike 0.0 0.0 \sparkspike 0.1111 0.0066 \sparkspike 0.2222 0.0773 \sparkspike 0.3333 0.521 \sparkspike 0.4444 0.9073 \sparkspike 0.5556 1.0 \sparkspike 0.6667 0.468 \sparkspike 0.7778 0.223 \sparkspike 0.8889 0.0265 \sparkspike 1.0 0.0044 \end{sparkline}
& \scriptsize{0.32} & \scriptsize{0.06} & 
\begin{sparkline}{10} \definecolor{sparkbottomlinecolor}{gray}{0.8} \setlength\sparkbottomlinethickness{0.7pt} \sparkbottomlinex -0.05 1.05 \sparkspike 0.0 0.0024 \sparkspike 0.1111 0.0694 \sparkspike 0.2222 0.5104 \sparkspike 0.3333 1.0 \sparkspike 0.4444 0.1973 \sparkspike 0.5556 0.0049 \sparkspike 0.6667 0.0 \sparkspike 0.7778 0.0 \sparkspike 0.8889 0.0 \sparkspike 1.0 0.0 \end{sparkline}
\\
\scriptsize{ANPST}~\cite{dataset_polish_sentences} & \scriptsize{Polish} & \scriptsize{718} & \scriptsize{28.16} & \scriptsize{0.48} & \scriptsize{0.13} & 
\begin{sparkline}{10} \definecolor{sparkbottomlinecolor}{gray}{0.8} \setlength\sparkbottomlinethickness{0.7pt} \sparkbottomlinex -0.05 1.05 \sparkspike 0.0 0.0 \sparkspike 0.1111 0.0619 \sparkspike 0.2222 0.2143 \sparkspike 0.3333 0.6476 \sparkspike 0.4444 1.0 \sparkspike 0.5556 0.8333 \sparkspike 0.6667 0.4381 \sparkspike 0.7778 0.1857 \sparkspike 0.8889 0.0381 \sparkspike 1.0 0.0 \end{sparkline}
& \scriptsize{0.47} & \scriptsize{0.22} & 
\begin{sparkline}{10} \definecolor{sparkbottomlinecolor}{gray}{0.8} \setlength\sparkbottomlinethickness{0.7pt} \sparkbottomlinex -0.05 1.05 \sparkspike 0.0 0.0588 \sparkspike 0.1111 0.7143 \sparkspike 0.2222 1.0 \sparkspike 0.3333 0.8571 \sparkspike 0.4444 0.7395 \sparkspike 0.5556 0.7479 \sparkspike 0.6667 0.6303 \sparkspike 0.7778 0.8739 \sparkspike 0.8889 0.4118 \sparkspike 1.0 0.0  \end{sparkline}
\\
\scriptsize{MAS}~\cite{dataset_mas} & \scriptsize{Portuguese} & \scriptsize{192} & \scriptsize{8.94} & \scriptsize{0.52} & \scriptsize{0.17} & 
\begin{sparkline}{10} \definecolor{sparkbottomlinecolor}{gray}{0.8} \setlength\sparkbottomlinethickness{0.7pt} \sparkbottomlinex -0.05 1.05 \sparkspike 0.0 0.0 \sparkspike 0.1111 0.0 \sparkspike 0.2222 0.0645 \sparkspike 0.3333 1.0 \sparkspike 0.4444 0.629 \sparkspike 0.5556 0.3548 \sparkspike 0.6667 0.2903 \sparkspike 0.7778 0.5161 \sparkspike 0.8889 0.2419 \sparkspike 1.0 0.0 \end{sparkline}
& \scriptsize{0.49} & \scriptsize{0.28} & 
\begin{sparkline}{10} \definecolor{sparkbottomlinecolor}{gray}{0.8} \setlength\sparkbottomlinethickness{0.7pt} \sparkbottomlinex -0.05 1.05 \sparkspike 0.0 0.5 \sparkspike 0.1111 0.8095 \sparkspike 0.2222 0.1429 \sparkspike 0.3333 0.0 \sparkspike 0.4444 0.4286 \sparkspike 0.5556 1.0 \sparkspike 0.6667 0.0714 \sparkspike 0.7778 0.619 \sparkspike 0.8889 0.6429 \sparkspike 1.0 0.119 \end{sparkline}
\\
 \midrule

\scriptsize{Yee}~\cite{dataset_cantonese_nouns} & \scriptsize{Cantonese} & \scriptsize{292} & & \scriptsize{0.40} & \scriptsize{0.15} & 
\begin{sparkline}{10} \definecolor{sparkbottomlinecolor}{gray}{0.8} \setlength\sparkbottomlinethickness{0.7pt} \sparkbottomlinex -0.05 1.05 \sparkspike 0.0 0.0 \sparkspike 0.1111 0.1807 \sparkspike 0.2222 0.9036 \sparkspike 0.3333 1.0 \sparkspike 0.4444 0.6627 \sparkspike 0.5556 0.4217 \sparkspike 0.6667 0.1446 \sparkspike 0.7778 0.1446 \sparkspike 0.8889 0.0602 \sparkspike 1.0 0.0 \end{sparkline}
& \scriptsize{0.58} & \scriptsize{0.17} & 
\begin{sparkline}{10} \definecolor{sparkbottomlinecolor}{gray}{0.8} \setlength\sparkbottomlinethickness{0.7pt} \sparkbottomlinex -0.05 1.05 \sparkspike 0.0 0.0 \sparkspike 0.1111 0.0682 \sparkspike 0.2222 0.0568 \sparkspike 0.3333 0.1136 \sparkspike 0.4444 0.4432 \sparkspike 0.5556 1.0 \sparkspike 0.6667 0.6023 \sparkspike 0.7778 0.375 \sparkspike 0.8889 0.2841 \sparkspike 1.0 0.0341 \end{sparkline}
\\
\scriptsize{Ćoso et al.}~\cite{dataset_croatian_norms} & \scriptsize{Croatian} & \scriptsize{3022} & & \scriptsize{0.45} & \scriptsize{0.15} & 
\begin{sparkline}{10} \definecolor{sparkbottomlinecolor}{gray}{0.8} \setlength\sparkbottomlinethickness{0.7pt} \sparkbottomlinex -0.05 1.05 \sparkspike 0.0 0.0014 \sparkspike 0.1111 0.0962 \sparkspike 0.2222 0.6099 \sparkspike 0.3333 1.0 \sparkspike 0.4444 0.9162 \sparkspike 0.5556 0.7527 \sparkspike 0.6667 0.4808 \sparkspike 0.7778 0.2184 \sparkspike 0.8889 0.0481 \sparkspike 1.0 0.0 \end{sparkline}
& \scriptsize{0.51} & \scriptsize{0.21} & 
\begin{sparkline}{10} \definecolor{sparkbottomlinecolor}{gray}{0.8} \setlength\sparkbottomlinethickness{0.7pt} \sparkbottomlinex -0.05 1.05 \sparkspike 0.0 0.0505 \sparkspike 0.1111 0.4218 \sparkspike 0.2222 0.4625 \sparkspike 0.3333 0.3567 \sparkspike 0.4444 0.5358 \sparkspike 0.5556 1.0 \sparkspike 0.6667 0.8453 \sparkspike 0.7778 0.5635 \sparkspike 0.8889 0.2785 \sparkspike 1.0 0.0 \end{sparkline}
\\
\scriptsize{Moors et al.}~\cite{dataset_dutch_norms} & \scriptsize{Dutch} & \scriptsize{4299} & & \scriptsize{0.52} & \scriptsize{0.14} & 
\begin{sparkline}{10} \definecolor{sparkbottomlinecolor}{gray}{0.8} \setlength\sparkbottomlinethickness{0.7pt} \sparkbottomlinex -0.05 1.05 \sparkspike 0.0 0.0 \sparkspike 0.1111 0.02 \sparkspike 0.2222 0.1146 \sparkspike 0.3333 0.7616 \sparkspike 0.4444 1.0 \sparkspike 0.5556 0.9245 \sparkspike 0.6667 0.6278 \sparkspike 0.7778 0.3276 \sparkspike 0.8889 0.1137 \sparkspike 1.0 0.0036 \end{sparkline}
& \scriptsize{0.49} & \scriptsize{0.18} & 
\begin{sparkline}{10} \definecolor{sparkbottomlinecolor}{gray}{0.8} \setlength\sparkbottomlinethickness{0.7pt} \sparkbottomlinex -0.05 1.05 \sparkspike 0.0 0.0231 \sparkspike 0.1111 0.1742 \sparkspike 0.2222 0.4261 \sparkspike 0.3333 0.422 \sparkspike 0.4444 0.5681 \sparkspike 0.5556 1.0 \sparkspike 0.6667 0.4624 \sparkspike 0.7778 0.3105 \sparkspike 0.8889 0.114 \sparkspike 1.0 0.0 \end{sparkline}
\\
\scriptsize{Verheyen et al.}~\cite{dataset_dutch_adjectives} & \scriptsize{Dutch} & \scriptsize{1000} & & \scriptsize{0.52} & \scriptsize{0.17} & 
\begin{sparkline}{10} \definecolor{sparkbottomlinecolor}{gray}{0.8} \setlength\sparkbottomlinethickness{0.7pt} \sparkbottomlinex -0.05 1.05 \sparkspike 0.0 0.0 \sparkspike 0.1111 0.1066 \sparkspike 0.2222 0.3279 \sparkspike 0.3333 0.668 \sparkspike 0.4444 0.7992 \sparkspike 0.5556 1.0 \sparkspike 0.6667 0.5205 \sparkspike 0.7778 0.4795 \sparkspike 0.8889 0.1803 \sparkspike 1.0 0.0164 \end{sparkline}
& \scriptsize{0.50} & \scriptsize{0.20} & 
\begin{sparkline}{10} \definecolor{sparkbottomlinecolor}{gray}{0.8} \setlength\sparkbottomlinethickness{0.7pt} \sparkbottomlinex -0.05 1.05 \sparkspike 0.0 0.0 \sparkspike 0.1111 0.3086 \sparkspike 0.2222 0.8642 \sparkspike 0.3333 0.8148 \sparkspike 0.4444 0.8272 \sparkspike 0.5556 1.0 \sparkspike 0.6667 0.858 \sparkspike 0.7778 0.8086 \sparkspike 0.8889 0.3148 \sparkspike 1.0 0.0062 \end{sparkline}
\\
\scriptsize{NRC-VAD}~\cite{dataset_nrc-vad} & \scriptsize{English} & \scriptsize{19971} & & \scriptsize{0.50} & \scriptsize{0.17} & 
\begin{sparkline}{10} \definecolor{sparkbottomlinecolor}{gray}{0.8} \setlength\sparkbottomlinethickness{0.7pt} \sparkbottomlinex -0.05 1.05 \sparkspike 0.0 0.0 \sparkspike 0.1111 0.0727 \sparkspike 0.2222 0.3846 \sparkspike 0.3333 0.8113 \sparkspike 0.4444 1.0 \sparkspike 0.5556 0.8168 \sparkspike 0.6667 0.5278 \sparkspike 0.7778 0.3332 \sparkspike 0.8889 0.1953 \sparkspike 1.0 0.0574 \end{sparkline}
& \scriptsize{0.50} & \scriptsize{0.22} & 
\begin{sparkline}{10} \definecolor{sparkbottomlinecolor}{gray}{0.8} \setlength\sparkbottomlinethickness{0.7pt} \sparkbottomlinex -0.05 1.05 \sparkspike 0.0 0.0443 \sparkspike 0.1111 0.3251 \sparkspike 0.2222 0.3673 \sparkspike 0.3333 0.506 \sparkspike 0.4444 0.7804 \sparkspike 0.5556 1.0 \sparkspike 0.6667 0.7426 \sparkspike 0.7778 0.4277 \sparkspike 0.8889 0.2219 \sparkspike 1.0 0.0 \end{sparkline}
\\
\scriptsize{Warriner et al.}~\cite{dataset_english_norms} & \scriptsize{English} & \scriptsize{13915} & & \scriptsize{0.40} & \scriptsize{0.11} & 
\begin{sparkline}{10} \definecolor{sparkbottomlinecolor}{gray}{0.8} \setlength\sparkbottomlinethickness{0.7pt} \sparkbottomlinex -0.05 1.05 \sparkspike 0.0 0.0008 \sparkspike 0.1111 0.0417 \sparkspike 0.2222 0.5034 \sparkspike 0.3333 1.0 \sparkspike 0.4444 0.7965 \sparkspike 0.5556 0.3944 \sparkspike 0.6667 0.1397 \sparkspike 0.7778 0.024 \sparkspike 0.8889 0.0015 \sparkspike 1.0 0.0 \end{sparkline}
& \scriptsize{0.51} & \scriptsize{0.16} & 
\begin{sparkline}{10} \definecolor{sparkbottomlinecolor}{gray}{0.8} \setlength\sparkbottomlinethickness{0.7pt} \sparkbottomlinex -0.05 1.05 \sparkspike 0.0 0.0069 \sparkspike 0.1111 0.1156 \sparkspike 0.2222 0.309 \sparkspike 0.3333 0.4064 \sparkspike 0.4444 0.6551 \sparkspike 0.5556 1.0 \sparkspike 0.6667 0.6394 \sparkspike 0.7778 0.2968 \sparkspike 0.8889 0.0714 \sparkspike 1.0 0.0 \end{sparkline}
\\
\scriptsize{Scott et al.}~\cite{dataset_glasgow_norms} & \scriptsize{English} & \scriptsize{5553} & & \scriptsize{0.45} & \scriptsize{0.14} & 
\begin{sparkline}{10} \definecolor{sparkbottomlinecolor}{gray}{0.8} \setlength\sparkbottomlinethickness{0.7pt} \sparkbottomlinex -0.05 1.05 \sparkspike 0.0 0.0226 \sparkspike 0.1111 0.209 \sparkspike 0.2222 0.2372 \sparkspike 0.3333 0.2341 \sparkspike 0.4444 0.4187 \sparkspike 0.5556 1.0 \sparkspike 0.6667 0.4963 \sparkspike 0.7778 0.2524 \sparkspike 0.8889 0.1571 \sparkspike 1.0 0.0 \end{sparkline}
& \scriptsize{0.51} & \scriptsize{0.19} & 
\begin{sparkline}{10} \definecolor{sparkbottomlinecolor}{gray}{0.8} \setlength\sparkbottomlinethickness{0.7pt} \sparkbottomlinex -0.05 1.05 \sparkspike 0.0 0.0 \sparkspike 0.1111 0.0304 \sparkspike 0.2222 0.4578 \sparkspike 0.3333 1.0 \sparkspike 0.4444 0.9426 \sparkspike 0.5556 0.7387 \sparkspike 0.6667 0.3883 \sparkspike 0.7778 0.1533 \sparkspike 0.8889 0.0385 \sparkspike 1.0 0.0 \end{sparkline}
\\
\scriptsize{Söderholm et al.}~\cite{dataset_finnish_nouns} & \scriptsize{Finnish} & \scriptsize{420} & & \scriptsize{0.50} & \scriptsize{0.13} & 
\begin{sparkline}{10} \definecolor{sparkbottomlinecolor}{gray}{0.8} \setlength\sparkbottomlinethickness{0.7pt} \sparkbottomlinex -0.05 1.05 \sparkspike 0.0 0.0 \sparkspike 0.1111 0.0216 \sparkspike 0.2222 0.0935 \sparkspike 0.3333 0.6547 \sparkspike 0.4444 1.0 \sparkspike 0.5556 0.5683 \sparkspike 0.6667 0.4245 \sparkspike 0.7778 0.223 \sparkspike 0.8889 0.036 \sparkspike 1.0 0.0 \end{sparkline}
& 0.50 & 0.25 & 
\begin{sparkline}{10} \definecolor{sparkbottomlinecolor}{gray}{0.8} \setlength\sparkbottomlinethickness{0.7pt} \sparkbottomlinex -0.05 1.05 \sparkspike 0.0 0.3188 \sparkspike 0.1111 0.6957 \sparkspike 0.2222 0.6667 \sparkspike 0.3333 0.3188 \sparkspike 0.4444 0.4203 \sparkspike 0.5556 0.913 \sparkspike 0.6667 1.0 \sparkspike 0.7778 0.9275 \sparkspike 0.8889 0.5362 \sparkspike 1.0 0.0 \end{sparkline}
\\
\scriptsize{Eilola et al.}~\cite{dataset_finnish_norms} & \scriptsize{Finnish} & \scriptsize{210} & & \scriptsize{0.36} & \scriptsize{0.19} & 
\begin{sparkline}{10} \definecolor{sparkbottomlinecolor}{gray}{0.8} \setlength\sparkbottomlinethickness{0.7pt} \sparkbottomlinex -0.05 1.05 \sparkspike 0.0 0.3902 \sparkspike 0.1111 1.0 \sparkspike 0.2222 0.8049 \sparkspike 0.3333 0.5854 \sparkspike 0.4444 0.878 \sparkspike 0.5556 0.6829 \sparkspike 0.6667 0.5854 \sparkspike 0.7778 0.1707 \sparkspike 0.8889 0.0244 \sparkspike 1.0 0.0 \end{sparkline}
& \scriptsize{0.44} & \scriptsize{0.26} & 
\begin{sparkline}{10} \definecolor{sparkbottomlinecolor}{gray}{0.8} \setlength\sparkbottomlinethickness{0.7pt} \sparkbottomlinex -0.05 1.05 \sparkspike 0.0 0.1321 \sparkspike 0.1111 1.0 \sparkspike 0.2222 0.4717 \sparkspike 0.3333 0.0189 \sparkspike 0.4444 0.6604 \sparkspike 0.5556 0.4906 \sparkspike 0.6667 0.283 \sparkspike 0.7778 0.5094 \sparkspike 0.8889 0.3962 \sparkspike 1.0 0.0 \end{sparkline}
\\
\scriptsize{FAN}~\cite{dataset_fan} & \scriptsize{French} & \scriptsize{1031} & & \scriptsize{0.41} & \scriptsize{0.13} & 
\begin{sparkline}{10} \definecolor{sparkbottomlinecolor}{gray}{0.8} \setlength\sparkbottomlinethickness{0.7pt} \sparkbottomlinex -0.05 1.05 \sparkspike 0.0 0.0 \sparkspike 0.1111 0.0741 \sparkspike 0.2222 0.6235 \sparkspike 0.3333 1.0 \sparkspike 0.4444 0.7006 \sparkspike 0.5556 0.4568 \sparkspike 0.6667 0.216 \sparkspike 0.7778 0.0957 \sparkspike 0.8889 0.0154 \sparkspike 1.0 0.0 \end{sparkline}
& \scriptsize{0.56} & \scriptsize{0.17} & 
\begin{sparkline}{10} \definecolor{sparkbottomlinecolor}{gray}{0.8} \setlength\sparkbottomlinethickness{0.7pt} \sparkbottomlinex -0.05 1.05 \sparkspike 0.0 0.0 \sparkspike 0.1111 0.0854 \sparkspike 0.2222 0.1179 \sparkspike 0.3333 0.2602 \sparkspike 0.4444 0.7683 \sparkspike 0.5556 1.0 \sparkspike 0.6667 0.7114 \sparkspike 0.7778 0.5569 \sparkspike 0.8889 0.1951 \sparkspike 1.0 0.0081 \end{sparkline}
\\
\scriptsize{FEEL}~\cite{dataset_feel} & \scriptsize{French} & \scriptsize{835} & & \scriptsize{0.56} & \scriptsize{0.17} & 
\begin{sparkline}{10} \definecolor{sparkbottomlinecolor}{gray}{0.8} \setlength\sparkbottomlinethickness{0.7pt} \sparkbottomlinex -0.05 1.05 \sparkspike 0.0 0.0 \sparkspike 0.1111 0.0688 \sparkspike 0.2222 0.3807 \sparkspike 0.3333 0.367 \sparkspike 0.4444 0.422 \sparkspike 0.5556 0.7982 \sparkspike 0.6667 1.0 \sparkspike 0.7778 0.6606 \sparkspike 0.8889 0.133 \sparkspike 1.0 0.0 \end{sparkline}
& \scriptsize{0.43} & \scriptsize{0.20} & 
\begin{sparkline}{10} \definecolor{sparkbottomlinecolor}{gray}{0.8} \setlength\sparkbottomlinethickness{0.7pt} \sparkbottomlinex -0.05 1.05 \sparkspike 0.0 0.0091 \sparkspike 0.1111 0.3545 \sparkspike 0.2222 1.0 \sparkspike 0.3333 0.7045 \sparkspike 0.4444 0.3227 \sparkspike 0.5556 0.3227 \sparkspike 0.6667 0.5591 \sparkspike 0.7778 0.4364 \sparkspike 0.8889 0.0409 \sparkspike 1.0 0.0 \end{sparkline}
\\
\scriptsize{BAWL-R}~\cite{dataset_bawl_r} & \scriptsize{German} & \scriptsize{2902} & & \scriptsize{0.44} & \scriptsize{0.17} & 
\begin{sparkline}{10} \definecolor{sparkbottomlinecolor}{gray}{0.8} \setlength\sparkbottomlinethickness{0.7pt} \sparkbottomlinex -0.05 1.05 \sparkspike 0.0 0.0131 \sparkspike 0.1111 0.1985 \sparkspike 0.2222 0.7109 \sparkspike 0.3333 1.0 \sparkspike 0.4444 0.7898 \sparkspike 0.5556 0.6423 \sparkspike 0.6667 0.4686 \sparkspike 0.7778 0.2423 \sparkspike 0.8889 0.0832 \sparkspike 1.0 0.0 \end{sparkline}
& \scriptsize{0.51} & \scriptsize{0.21} & 
\begin{sparkline}{10} \definecolor{sparkbottomlinecolor}{gray}{0.8} \setlength\sparkbottomlinethickness{0.7pt} \sparkbottomlinex -0.05 1.05 \sparkspike 0.0 0.0058 \sparkspike 0.1111 0.3779 \sparkspike 0.2222 0.5581 \sparkspike 0.3333 0.6008 \sparkspike 0.4444 0.5523 \sparkspike 0.5556 1.0 \sparkspike 0.6667 0.8314 \sparkspike 0.7778 0.7519 \sparkspike 0.8889 0.2287 \sparkspike 1.0 0.0 \end{sparkline}
\\
\scriptsize{ANGST}~\cite{dataset_angst} & \scriptsize{German} & \scriptsize{1034} & & \scriptsize{0.52} & \scriptsize{0.16} & 
\begin{sparkline}{10} \definecolor{sparkbottomlinecolor}{gray}{0.8} \setlength\sparkbottomlinethickness{0.7pt} \sparkbottomlinex -0.05 1.05 \sparkspike 0.0 0.0 \sparkspike 0.1111 0.019 \sparkspike 0.2222 0.381 \sparkspike 0.3333 0.881 \sparkspike 0.4444 0.9905 \sparkspike 0.5556 1.0 \sparkspike 0.6667 0.9476 \sparkspike 0.7778 0.5762 \sparkspike 0.8889 0.1286 \sparkspike 1.0 0.0 \end{sparkline}
& \scriptsize{0.51} & \scriptsize{0.24} & 
\begin{sparkline}{10} \definecolor{sparkbottomlinecolor}{gray}{0.8} \setlength\sparkbottomlinethickness{0.7pt} \sparkbottomlinex -0.05 1.05 \sparkspike 0.0 0.0506 \sparkspike 0.1111 0.7089 \sparkspike 0.2222 0.6962 \sparkspike 0.3333 0.4367 \sparkspike 0.4444 0.2532 \sparkspike 0.5556 1.0 \sparkspike 0.6667 0.7215 \sparkspike 0.7778 0.8987 \sparkspike 0.8889 0.5127 \sparkspike 1.0 0.0 \end{sparkline}
\\
\scriptsize{LANG}~\cite{dataset_leipzig_norms} & \scriptsize{German} & \scriptsize{1000} & & \scriptsize{0.39} & \scriptsize{0.20} & 
\begin{sparkline}{10} \definecolor{sparkbottomlinecolor}{gray}{0.8} \setlength\sparkbottomlinethickness{0.7pt} \sparkbottomlinex -0.05 1.05 \sparkspike 0.0 0.0282 \sparkspike 0.1111 1.0 \sparkspike 0.2222 0.5927 \sparkspike 0.3333 0.5121 \sparkspike 0.4444 0.4919 \sparkspike 0.5556 0.625 \sparkspike 0.6667 0.504 \sparkspike 0.7778 0.246 \sparkspike 0.8889 0.0323 \sparkspike 1.0 0.0 \end{sparkline}
& \scriptsize{0.50} & \scriptsize{0.13} & 
\begin{sparkline}{10} \definecolor{sparkbottomlinecolor}{gray}{0.8} \setlength\sparkbottomlinethickness{0.7pt} \sparkbottomlinex -0.05 1.05 \sparkspike 0.0 0.0 \sparkspike 0.1111 0.0081 \sparkspike 0.2222 0.2769 \sparkspike 0.3333 0.4247 \sparkspike 0.4444 0.3817 \sparkspike 0.5556 1.0 \sparkspike 0.6667 0.4919 \sparkspike 0.7778 0.1048 \sparkspike 0.8889 0.0 \sparkspike 1.0 0.0 \end{sparkline}
\\
\scriptsize{Italian ANEW}~\cite{dataset_italian_words} & \scriptsize{Italian} & \scriptsize{1121} & & \scriptsize{0.52} & \scriptsize{0.19} & 
\begin{sparkline}{10} \definecolor{sparkbottomlinecolor}{gray}{0.8} \setlength\sparkbottomlinethickness{0.7pt} \sparkbottomlinex -0.05 1.05 \sparkspike 0.0 0.0 \sparkspike 0.1111 0.1031 \sparkspike 0.2222 0.5471 \sparkspike 0.3333 1.0 \sparkspike 0.4444 0.8655 \sparkspike 0.5556 0.6682 \sparkspike 0.6667 0.6547 \sparkspike 0.7778 0.7937 \sparkspike 0.8889 0.3408 \sparkspike 1.0 0.009 \end{sparkline}
& \scriptsize{0.51} & \scriptsize{0.26} & 
\begin{sparkline}{10} \definecolor{sparkbottomlinecolor}{gray}{0.8} \setlength\sparkbottomlinethickness{0.7pt} \sparkbottomlinex -0.05 1.05 \sparkspike 0.0 0.0872 \sparkspike 0.1111 0.9282 \sparkspike 0.2222 0.5128 \sparkspike 0.3333 0.241 \sparkspike 0.4444 0.2205 \sparkspike 0.5556 0.5128 \sparkspike 0.6667 0.8308 \sparkspike 0.7778 1.0 \sparkspike 0.8889 0.5436 \sparkspike 1.0 0.0 \end{sparkline}
\\
\scriptsize{Xu et al.}~\cite{dataset_chinese_words} & \scriptsize{Mandarin} & \scriptsize{11310} & & \scriptsize{0.52} & \scriptsize{0.14} & 
\begin{sparkline}{10} \definecolor{sparkbottomlinecolor}{gray}{0.8} \setlength\sparkbottomlinethickness{0.7pt} \sparkbottomlinex -0.05 1.05 \sparkspike 0.0 0.0 \sparkspike 0.1111 0.007 \sparkspike 0.2222 0.1516 \sparkspike 0.3333 0.6134 \sparkspike 0.4444 0.9202 \sparkspike 0.5556 1.0 \sparkspike 0.6667 0.6646 \sparkspike 0.7778 0.3145 \sparkspike 0.8889 0.0841 \sparkspike 1.0 0.0047 \end{sparkline}
& \scriptsize{0.52} & \scriptsize{0.16} & 
\begin{sparkline}{10} \definecolor{sparkbottomlinecolor}{gray}{0.8} \setlength\sparkbottomlinethickness{0.7pt} \sparkbottomlinex -0.05 1.05 \sparkspike 0.0 0.0113 \sparkspike 0.1111 0.098 \sparkspike 0.2222 0.19 \sparkspike 0.3333 0.2008 \sparkspike 0.4444 0.297 \sparkspike 0.5556 1.0 \sparkspike 0.6667 0.3795 \sparkspike 0.7778 0.1997 \sparkspike 0.8889 0.0741 \sparkspike 1.0 0.0 \end{sparkline}
\\
\scriptsize{CVAW}~\cite{dataset_chinese_emobank,dataset_cvaw_cvat} & \scriptsize{Mandarin} & \scriptsize{5512} & & \scriptsize{0.50} & \scriptsize{0.18} & 
\begin{sparkline}{10} \definecolor{sparkbottomlinecolor}{gray}{0.8} \setlength\sparkbottomlinethickness{0.7pt} \sparkbottomlinex -0.05 1.05 \sparkspike 0.0 0.0 \sparkspike 0.1111 0.0918 \sparkspike 0.2222 0.5008 \sparkspike 0.3333 0.659 \sparkspike 0.4444 0.8951 \sparkspike 0.5556 1.0 \sparkspike 0.6667 0.6033 \sparkspike 0.7778 0.3369 \sparkspike 0.8889 0.2443 \sparkspike 1.0 0.023 \end{sparkline}
& \scriptsize{0.44} & \scriptsize{0.21} & 
\begin{sparkline}{10} \definecolor{sparkbottomlinecolor}{gray}{0.8} \setlength\sparkbottomlinethickness{0.7pt} \sparkbottomlinex -0.05 1.05 \sparkspike 0.0 0.0335 \sparkspike 0.1111 0.2662 \sparkspike 0.2222 1.0 \sparkspike 0.3333 0.71 \sparkspike 0.4444 0.4119 \sparkspike 0.5556 0.452 \sparkspike 0.6667 0.5465 \sparkspike 0.7778 0.3613 \sparkspike 0.8889 0.1532 \sparkspike 1.0 0.0 \end{sparkline}
\\
\scriptsize{ANPW\_R}~\cite{dataset_anpw_r} & \scriptsize{Polish} & \scriptsize{4905} & & \scriptsize{0.39} & \scriptsize{0.11} & 
\begin{sparkline}{10} \definecolor{sparkbottomlinecolor}{gray}{0.8} \setlength\sparkbottomlinethickness{0.7pt} \sparkbottomlinex -0.05 1.05 \sparkspike 0.0 0.0 \sparkspike 0.1111 0.016 \sparkspike 0.2222 0.6193 \sparkspike 0.3333 1.0 \sparkspike 0.4444 0.6514 \sparkspike 0.5556 0.3204 \sparkspike 0.6667 0.0891 \sparkspike 0.7778 0.0183 \sparkspike 0.8889 0.0 \sparkspike 1.0 0.0 \end{sparkline}
& \scriptsize{0.50} & \scriptsize{0.16} & 
\begin{sparkline}{10} \definecolor{sparkbottomlinecolor}{gray}{0.8} \setlength\sparkbottomlinethickness{0.7pt} \sparkbottomlinex -0.05 1.05 \sparkspike 0.0 0.006 \sparkspike 0.1111 0.1027 \sparkspike 0.2222 0.396 \sparkspike 0.3333 0.3705 \sparkspike 0.4444 0.4953 \sparkspike 0.5556 1.0 \sparkspike 0.6667 0.5792 \sparkspike 0.7778 0.2725 \sparkspike 0.8889 0.0564 \sparkspike 1.0 0.0 \end{sparkline}
\\
\scriptsize{NAWL}~\cite{dataset_nawl} & \scriptsize{Polish} & \scriptsize{2902} & & \scriptsize{0.34} & \scriptsize{0.13} & 
\begin{sparkline}{10} \definecolor{sparkbottomlinecolor}{gray}{0.8} \setlength\sparkbottomlinethickness{0.7pt} \sparkbottomlinex -0.05 1.05 \sparkspike 0.0 0.0606 \sparkspike 0.1111 0.5316 \sparkspike 0.2222 0.9919 \sparkspike 0.3333 1.0 \sparkspike 0.4444 0.7537 \sparkspike 0.5556 0.4361 \sparkspike 0.6667 0.1184 \sparkspike 0.7778 0.0121 \sparkspike 0.8889 0.0013 \sparkspike 1.0 0.0 \end{sparkline}
& \scriptsize{0.53} & \scriptsize{0.20} & 
\begin{sparkline}{10} \definecolor{sparkbottomlinecolor}{gray}{0.8} \setlength\sparkbottomlinethickness{0.7pt} \sparkbottomlinex -0.05 1.05 \sparkspike 0.0 0.0 \sparkspike 0.1111 0.2597 \sparkspike 0.2222 0.4699 \sparkspike 0.3333 0.4235 \sparkspike 0.4444 0.4791 \sparkspike 0.5556 1.0 \sparkspike 0.6667 0.7635 \sparkspike 0.7778 0.6213 \sparkspike 0.8889 0.2767 \sparkspike 1.0 0.0216 \end{sparkline}
\\
\scriptsize{Portuguese ANEW}~\cite{dataset_anew_ep} & \scriptsize{Portuguese} & \scriptsize{1034} & & \scriptsize{0.49} & \scriptsize{0.14} & 
\begin{sparkline}{10} \definecolor{sparkbottomlinecolor}{gray}{0.8} \setlength\sparkbottomlinethickness{0.7pt} \sparkbottomlinex -0.05 1.05 \sparkspike 0.0 0.0039 \sparkspike 0.1111 0.0502 \sparkspike 0.2222 0.2394 \sparkspike 0.3333 1.0 \sparkspike 0.4444 0.9112 \sparkspike 0.5556 0.8533 \sparkspike 0.6667 0.5985 \sparkspike 0.7778 0.3012 \sparkspike 0.8889 0.0347 \sparkspike 1.0 0.0 \end{sparkline}
& \scriptsize{0.50} & \scriptsize{0.23} & 
\begin{sparkline}{10} \definecolor{sparkbottomlinecolor}{gray}{0.8} \setlength\sparkbottomlinethickness{0.7pt} \sparkbottomlinex -0.05 1.05 \sparkspike 0.0 0.0113 \sparkspike 0.1111 0.5876 \sparkspike 0.2222 0.8475 \sparkspike 0.3333 0.4859 \sparkspike 0.4444 0.4124 \sparkspike 0.5556 1.0 \sparkspike 0.6667 0.7401 \sparkspike 0.7778 0.8079 \sparkspike 0.8889 0.4972 \sparkspike 1.0 0.0 \end{sparkline}
\\
\scriptsize{S.-Gonzalez et al.}~\cite{dataset_spanish_norms} & \scriptsize{Spanish} & \scriptsize{14031} & & \scriptsize{0.70} & \scriptsize{0.22} & 
\begin{sparkline}{10} \definecolor{sparkbottomlinecolor}{gray}{0.8} \setlength\sparkbottomlinethickness{0.7pt} \sparkbottomlinex -0.05 1.05 \sparkspike 0.0 0.0 \sparkspike 0.1111 0.0047 \sparkspike 0.2222 0.0136 \sparkspike 0.3333 0.0411 \sparkspike 0.4444 0.1277 \sparkspike 0.5556 0.5568 \sparkspike 0.6667 1.0 \sparkspike 0.7778 0.9168 \sparkspike 0.8889 0.4825 \sparkspike 1.0 0.3414 \end{sparkline}
& \scriptsize{0.72} & \scriptsize{0.16} & 
\begin{sparkline}{10} \definecolor{sparkbottomlinecolor}{gray}{0.8} \setlength\sparkbottomlinethickness{0.7pt} \sparkbottomlinex -0.05 1.05 \sparkspike 0.0 0.0 \sparkspike 0.1111 0.064 \sparkspike 0.2222 0.1804 \sparkspike 0.3333 0.207 \sparkspike 0.4444 0.2145 \sparkspike 0.5556 0.392 \sparkspike 0.6667 0.7517 \sparkspike 0.7778 1.0 \sparkspike 0.8889 0.5824 \sparkspike 1.0 0.389 \end{sparkline}
\\
\scriptsize{Kapucu et al.}~\cite{dataset_turkish_norms} & \scriptsize{Turkish} & \scriptsize{2031} & & \scriptsize{0.50} & \scriptsize{0.11} & 
\begin{sparkline}{10} \definecolor{sparkbottomlinecolor}{gray}{0.8} \setlength\sparkbottomlinethickness{0.7pt} \sparkbottomlinex -0.05 1.05 \sparkspike 0.0 0.0 \sparkspike 0.1111 0.0031 \sparkspike 0.2222 0.1207 \sparkspike 0.3333 0.5345 \sparkspike 0.4444 0.8777 \sparkspike 0.5556 1.0 \sparkspike 0.6667 0.5392 \sparkspike 0.7778 0.0956 \sparkspike 0.8889 0.0125 \sparkspike 1.0 0.0 \end{sparkline}
& \scriptsize{0.49} & \scriptsize{0.20} & 
\begin{sparkline}{10} \definecolor{sparkbottomlinecolor}{gray}{0.8} \setlength\sparkbottomlinethickness{0.7pt} \sparkbottomlinex -0.05 1.05 \sparkspike 0.0 0.0729 \sparkspike 0.1111 0.3794 \sparkspike 0.2222 0.2598 \sparkspike 0.3333 0.2897 \sparkspike 0.4444 0.7439 \sparkspike 0.5556 1.0 \sparkspike 0.6667 0.4262 \sparkspike 0.7778 0.372 \sparkspike 0.8889 0.1963 \sparkspike 1.0 0.0 \end{sparkline}
\\
 \bottomrule
\end{tabular}
\end{table*}

Overall, merging the 34 datasets allowed us to build a large multilingual VA dataset, consisting of 128,987 independently annotated instances~(i.e., 30,657 short texts and 98,330 words).
The textual sequences were left unchanged from the source datasets. As for the valence and arousal ratings, we took the mean annotated values when ratings were obtained from multiple individuals, and normalized the scores between 0 and 1. The normalization was performed according to the equation $z_i = (x_i - \textrm{min}(x)) / (\textrm{max}(x) - \textrm{min}(x))$, in which $z_i$ denotes the normalized value, $x_i$ the original value, and $\textrm{min}$ and $\textrm{max}$ denote the extremes of the scales in which the original scores were rated on.

\begin{figure}[htb!]
\centering
\includegraphics[width=.31\textwidth]{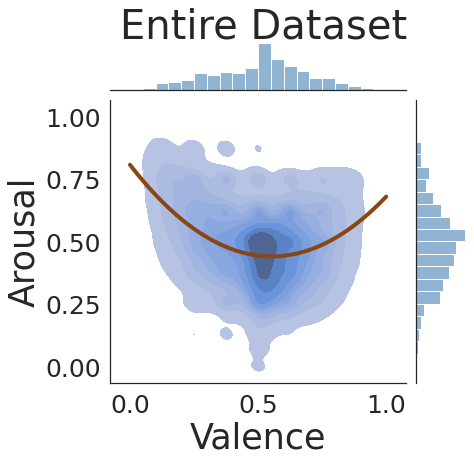}\hfill
\includegraphics[width=.31\textwidth]{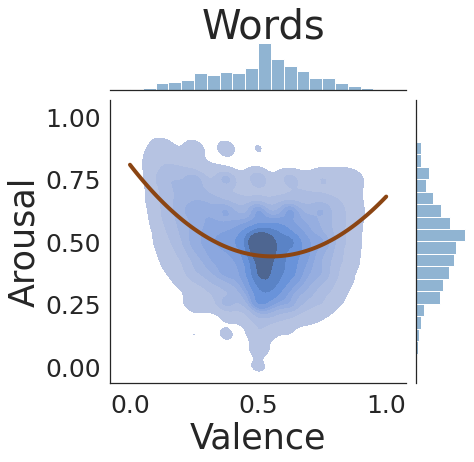}\hfill
\includegraphics[width=.31\textwidth]{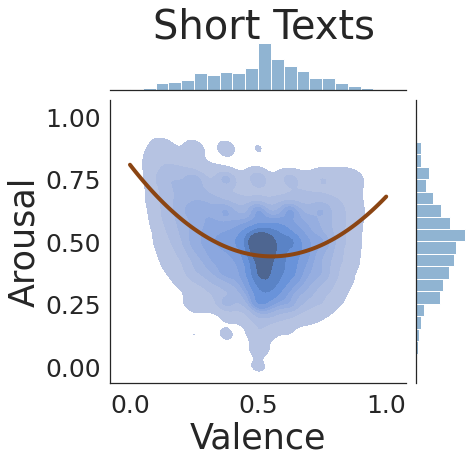}
\caption{Distribution of dataset instances in the valence-arousal space. Each dimensions' distribution is shown with a histogram on the corresponding axis. An orange trend line shows the quadratic relation between valence and arousal.}

\label{fig:data_distribution}
\end{figure}

Table~\ref{tab:ds_characterization} presents a statistical characterization for the short text datasets in its first half, followed by the word datasets. Each entry describes the dataset source language, the dataset size, and the mean number of words (this last variable in the case of the short texts). An exploratory analysis of the VA ratings supports a better understanding of the score distributions.
In turn, Figure~\ref{fig:data_distribution} presents the distribution of the ratings for the entire merged dataset, as well as for its two subsets (i.e., words and short texts). The ratings were plotted on the two-dimensional valence-arousal space, and they are visualized with the help of a kernel density estimate. The individual distributions of the two dimensions are displayed on the margins. The analysis of the resulting merged dataset leads to the conclusion that there is a quadratic relationship between the two emotional dimensions, with a tendency for increased arousal on high and low valence values, and abundant low arousal scores in the middle of the valence~scale. A similar pattern was previously observed in several different studies in Psychology, such as in the original ANEW study and its extensions~\cite{anew,dataset_croatian_norms,dataset_polish_sentences,kron_quadratic,dataset_fan,dataset_mas,dataset_chinese_words}.

\section{Experimental Evaluation}\label{section:exp_results}

Each of the individual original datasets were randomly split in half and combined with the others to form two subsets of data equally representative of all the datasets, later used for 2-fold cross-validation. 
For each configuration, two models were separately trained on each fold, and then separately used to make predictions for the instances in the other fold (containing instances not seen during training), with final evaluation metrics computed on the complete set of results (the predictions from the models trained on each fold were joined, and the metrics were computed over the full set of predictions).
Hyperparameters were defined through an initial set of tests and kept constant for all models. 
The batch size was fixed at 16, and models were trained during 10 epochs. 
AdamW was the chosen optimizer, and we used it together with a linear learning rate schedule with warm-up. The learning rate was set at $6\cdot10^{-6}$, with a warm-up ratio of~$1\cdot10^{-1}$.
We experimented with various model and loss function combinations, namely by using the three differently-sized pre-trained Transformer models, as well as the loss functions presented in Section \ref{section:models}. 

Three different evaluation metrics were used to assess and compare model performance, namely the Mean Absolute Error~(MAE), the Root Mean Squared Error~(RMSE), and the Pearson correlation coefficient~($\rho$). The MAE, as detailed by Equation~\ref{eqn: mae}, corresponds to the sum of absolute errors between observations $x_{i}$ and predictions $y_{i}$. 
\begin{equation}
  \mathrm{MAE} = \frac{1}{N}\sum_{i=1}^{N}|x_i-y_i|.
    \label{eqn: mae}
\end{equation}
The RMSE, as shown by Equation \ref{eqn: rmse}, is the square root of the mean square of the differences between observations $x_{i}$ and predictions $y_{i}$. 
\begin{equation}
    \mathrm{RMSE} = \sqrt{\frac{1}{N}\sum_{i=1}^{N}(x_i-y_i)^2}.
    \label{eqn: rmse}
\end{equation}
Finally, the Pearson correlation coefficient, given by Equation \ref{eqn: pearson correlation}, is used to assess the presence of a linear relationship between the ground truth $x$ and the predicted results given by $y$. 
\begin{equation}
    \rho = \frac{\sum_{i=1}^{N}(x_{i}-\bar{x})(y_{i}-\bar{y})}{\sqrt{\sum_{i=1}^{N}(x_{i}-\bar{x})^{2}(y_{i}-\bar{y})^{2}}}.
    \label{eqn: pearson correlation}
\end{equation}
While the first two metrics should be minimized, the latter is best when it is closer to one, i.e., the value denoting a perfect correlation.



\subsection{Results with Different Models and Loss Functions}
Table \ref{tab:general_results} summarizes the results for the different combinations of model size and loss function. 
The single thing that affects evaluation metrics the most is the size of the pre-trained Transformer model being used. The best performing model was the large version of XLM-RoBERTa, respectively returning on average 9\% and 20\% better correlation results than XLM-RoBERTa-base and DistilBERT. For each model, we compared the five loss functions, highlighting in bold the best performing one for each metric, and evaluating separately for valence and arousal. In short, the choice of loss function has less impact on the quality of the results. For the best model, we see differences in correlation of up to 2\% between best and worst performing loss functions. Comparatively, in the error metrics, these differences can be of up to 12\%. As such, looking to identify the best model/loss-function combination, we gave more relevance to the error metrics. We identified the MSE loss function as the best performing one, adding to the fact that this loss function is also the simplest of the set of functions that were tested. Consequently, further results are presented for that model/loss pair.

\begin{table}[tb!]\centering
\caption{Comparison between different models and loss functions.}
 \label{tab:general_results}
 \scriptsize
 \setlength{\tabcolsep}{5pt}
\begin{tabular}{cccccccc}
\hline
\footnotesize{\textbf{Model}}                                                                & \footnotesize{\textbf{Loss}} & \footnotesize{\boldmath$\mathrm{\rho_{V}}$} & \footnotesize{\boldmath$\mathrm{\rho_{A}}$} & \footnotesize{\boldmath$\mathrm{RMSE_{V}}$} & \footnotesize{\boldmath$\mathrm{RMSE_{A}}$} & \footnotesize{\boldmath$\mathrm{MAE_{V}}$} & \footnotesize{\boldmath$\mathrm{MAE_{A}}$} \\ \hline
  & MSE  & 0.663 & 0.594  & 0.138  & 0.132   & 0.102   & 0.101  \\
 & CCCL & 0.657 & 0.590 & 0.150 & 0.146 & 0.111 & 0.111 \\
 & RL & \textbf{0.668}  & \textbf{0.598}  & \textbf{0.138}     & \textbf{0.132} & \textbf{0.101} & \textbf{0.101}    \\
 & MSE+CCCL & 0.657 & 0.590 & 0.149 & 0.145 & 0.110 & 0.111 \\
\multirow{-5}{*}{DistilBERT} & RL+CCCL & 0.664 & 0.591 & 0.147 & 0.144 & 0.109 & 0.110 \\ \hline
 & MSE & 0.757 & 0.646 & \textbf{0.121}     & 0.125 & \textbf{0.088} & 0.095 \\
 & CCCL & 0.757 & 0.653 & 0.136 & 0.144 & 0.101 & 0.110 \\
 & RL & 0.757 & \textbf{0.657}  & 0.122 & \textbf{0.125} & 0.088 & \textbf{0.095}    \\
 & MSE+CCCL & 0.757 & 0.655 & 0.135 & 0.141 & 0.099 & 0.108 \\
\multirow{-5}{*}{\begin{tabular}[c]{@{}c@{}}XLM\\ RoBERTa\\ base\end{tabular}}  & RL+CCCL & \textbf{0.757}  & 0.657 & 0.134 & 0.141 & 0.099 & 0.107 \\ \hline
 & MSE & 0.810 & 0.695 & \textbf{0.109} & \textbf{0.120} & \textbf{0.079}    & \textbf{0.091}    \\
 & CCCL & \textbf{0.817}  & 0.698 & 0.117 & 0.132 & 0.085 & 0.099 \\
 & RL & 0.802 & 0.689 & 0.114 & 0.122 & 0.083 & 0.092 \\
 & MSE+CCCL & 0.815 & \textbf{0.699}  & 0.121 & 0.135 & 0.089 & 0.103 \\
\multirow{-5}{*}{\begin{tabular}[c]{@{}c@{}}XLM\\RoBERTa\\ large\end{tabular}} & RL+CCCL & 0.813 & 0.694 & 0.119 & 0.133 & 0.087 & 0.100 \\
\hline
\end{tabular}
\end{table}

When analyzing the results, it is possible to break them down into two categories: predicting valence and arousal for individual words or, on the other hand, for short texts (see Table \ref{tab:results_words_v_texts}). Our models are more accurate at predicting word-level scores, although this is also a more straightforward problem with less ambiguity. 
An essential fact to take from the results is the greater difficulty in predicting the affective dimension of arousal. Previous research has also stated that human ratings themselves varied much more in annotating arousal when compared to the valence dimension~\cite{Paltoglou}.

\begin{table}[tb!]\centering
\caption{Comparing VA prediction on words or short texts using the XLM-RoBERTa-large model and considering the MSE loss function for training.} \label{tab:results_words_v_texts}
\scriptsize
\setlength{\tabcolsep}{8.9pt}
\begin{tabular}{lcccccc}
\hline
\footnotesize{\textbf{Dataset}} & \footnotesize{\boldmath$\mathrm{\rho_{V}}$} & \footnotesize{\boldmath$\mathrm{\rho_{A}}$} & \footnotesize{\boldmath$\mathrm{RMSE_{V}}$} & \footnotesize{\boldmath$\mathrm{RMSE_{A}}$} & \footnotesize{\boldmath$\mathrm{MAE_{V}}$} & \footnotesize{\boldmath$\mathrm{MAE_{A}}$} \\ \hline
\textbf{All data} & 0.810 & 0.695 & 0.109 & 0.120 & 0.079 & 0.091 \\
\textbf{Words} & 0.833 & 0.686 & 0.107 & 0.116 & 0.078 & 0.090 \\
\textbf{Short texts} & 0.682 & 0.711 & 0.115 & 0.132 & 0.082 & 0.093 \\
\hline
\end{tabular}
\end{table}

\subsection{Results per Language and Dataset}
Further analysis focused on the results of predicting ratings for each of the original datasets, with results summarized on Table \ref{tab:results_datasets}. 

For most word datasets, compared in the bottom half of Table \ref{tab:results_datasets}, our best model performed to high standards, showing a correlation between predicted values and the ground truth of around $0.8$ for valence and $0.7$ for arousal. As a comparison, when evaluating correlation on Warriner's dataset \cite{dataset_english_norms}, our work achieved $\rho_{V}=0.84$ and $\rho_{A}=0.65$, while Hollis \cite{Hollis} achieved $\rho_{V}=0.80$ and $\rho_{A}=0.63$. Although good scores are observed for most datasets, we can also identify some outliers, like in the case of the dataset from Kapucu et al. \cite{dataset_turkish_norms}. 

As for the short text datasets, compared in the top half of Table \ref{tab:results_datasets}, performance varies more significantly, with an overall lower correlation and a higher error. A particular case is the COMETA stories dataset~\cite{dataset_cometa}, which shows a correlation close to zero. The COMETA dataset is a database of conceptual metaphors, in which half of the text instances contain metaphors while the other half corresponds to their literal counterparts.
The obtained results indicate that even the best model does not cope well with metaphorical phrasing. 
Comparing our model to the method from Preo{\c{t}}iuc-Pietro et al. \cite{dataset_facebook_posts}, the correlation values we obtained for the Facebook Posts dataset were $\rho_{V}=0.80$ and $\rho_{A}=0.78$, while they achieved $\rho_{V}=0.65$ and $\rho_{A}=0.85$ (i.e., we have better results for valence, and worse for arousal). In \cite{dataset_cvaw_cvat}, Yu et al. predict VA on the CVAT dataset using the ratings obtained for the CVAW words. They obtained correlation results of $\rho_{V}=0.54$ and $\rho_{A}=0.16$, while our approach obtained $\rho_{V}=0.89$ and $\rho_{A}=0.62$. In subsequent research, the same team tried to predict VA ratings with different neural network approaches, including a model based on BERT, for which they obtained $\rho_{V}=0.87$ and $\rho_{A}=0.58$ on the same dataset~\cite{dataset_chinese_emobank}. 

It should be noted that all previous comparisons against other studies are merely indicative, given that the experimental conditions (e.g., the data splits used for training and evaluation) were very different.

\begin{table}[b!]
\centering
\caption{Evaluation results for the short texts (top) and words (bottom) datasets, using the XLM-RoBERTa-large model and considering the MSE loss.}
\label{tab:results_datasets}
\setlength{\tabcolsep}{3.7pt} 
\scriptsize
\begin{tabular}{llcccccc}
\hline
\textbf{\footnotesize{Dataset}}           & \textbf{\footnotesize{Language}} & \footnotesize{\boldmath$\mathrm{\rho_{V}}$} & \footnotesize{\boldmath$\mathrm{\rho_{A}}$} & \footnotesize{\boldmath$\mathrm{RMSE_{V}}$} & \footnotesize{\boldmath$\mathrm{RMSE_{A}}$} & \footnotesize{\boldmath$\mathrm{MAE_{V}}$} & \footnotesize{\boldmath$\mathrm{MAE_{A}}$} \\ \hline
EmoBank & English & 0.736 & 0.440 & 0.061 & 0.071 & 0.044 & 0.052 \\
IEMOCAP & English & 0.469 & 0.656 & 0.159 & 0.173 & 0.126 & 0.132 \\
Facebook Posts   & English & 0.797 & 0.776 & 0.098 & 0.176 & 0.075 & 0.124 \\
EmoTales & English & 0.560 & 0.405 & 0.127 & 0.123 & 0.095 & 0.091 \\
ANET & English & 0.920 & 0.859 & 0.135 & 0.111 & 0.095 & 0.087 \\
PANIG & German & 0.597 & 0.563 & 0.181 & 0.111 & 0.137 & 0.085 \\
COMETA sent. & German & 0.853 & 0.598 & 0.103 & 0.120 & 0.074 & 0.096 \\
COMETA stories   & German & 0.072 & 0.042 & 0.254 & 0.160 & 0.206 & 0.130 \\
CVAT & Mandarin & 0.890 & 0.623 & 0.082 & 0.105 & 0.062 & 0.085 \\
CVAI & Mandarin & 0.517 & 0.720 & 0.068 & 0.089 & 0.053 & 0.071 \\
ANPST & Polish & 0.868 & 0.607 & 0.113 & 0.111 & 0.082 & 0.089 \\
MAS & Portuguese & 0.935 & 0.694 & 0.115 & 0.124 & 0.082 & 0.100 \\

\hline
Yee & Cantonese & 0.875 & 0.718 & 0.090 & 0.121 & 0.069 & 0.099 \\
Ćoso et al. & Croatian & 0.784 & 0.646 & 0.133 & 0.120 & 0.096 & 0.093 \\
Moors et al. & Dutch & 0.776 & 0.653 & 0.116 & 0.125 & 0.081 & 0.098 \\
Verheyen et al. & Dutch & 0.791 & 0.637 & 0.130 & 0.137 & 0.096 & 0.109 \\
NRC-VAD & English & 0.858 & 0.754 & 0.111 & 0.124 & 0.086 & 0.097 \\
Warriner et al. & English & 0.843 & 0.655 & 0.101 & 0.114 & 0.078 & 0.090 \\
Scott et al. & English & 0.884 & 0.636 & 0.095 & 0.117 & 0.067 & 0.092 \\
S\"{o}derholm et al. & Finnish & 0.645 & 0.492 & 0.188 & 0.138 & 0.147 & 0.109 \\
Eilola et al. & Finnish & 0.807 & 0.534 & 0.164 & 0.191 & 0.117 & 0.161 \\
FAN & French & 0.755 & 0.605 & 0.116 & 0.112 & 0.086 & 0.087 \\
FEEL & French & 0.823 & 0.664 & 0.131 & 0.131 & 0.096 & 0.103 \\
BAWL-R & German & 0.749 & 0.629 & 0.139 & 0.133 & 0.101 & 0.105 \\
ANGST & German & 0.837 & 0.738 & 0.135 & 0.117 & 0.092 & 0.089 \\
LANG & German & 0.802 & 0.696 & 0.100 & 0.144 & 0.074 & 0.115 \\
Italian ANEW & Italian & 0.846 & 0.644 & 0.138 & 0.148 & 0.099 & 0.118 \\
Xu et al. & Mandarin & 0.882 & 0.754 & 0.078 & 0.098 & 0.055 & 0.077 \\
CVAW & Mandarin & 0.904 & 0.666 & 0.094 & 0.136 & 0.071 & 0.108 \\
ANPW\_R & Polish & 0.846 & 0.689 & 0.093 & 0.088 & 0.065 & 0.069 \\
NAWL & Polish & 0.828 & 0.581 & 0.111 & 0.122 & 0.081 & 0.096 \\
Portuguese ANEW & Portuguese & 0.893 & 0.779 & 0.106 & 0.103 & 0.074 & 0.081 \\
S.-Gonzalez et al.  & Spanish & 0.808 & 0.689 & 0.100 & 0.095 & 0.074 & 0.072 \\
Kapucu et al. & Turkish & 0.571 & 0.373 & 0.165 & 0.127 & 0.125 & 0.101 \\
\hline
\end{tabular}
\end{table}

We performed a similar comparison to evaluate the result quality in distinct languages, grouping prediction results by language. 
It was possible to conclude that our best model yields good results in most languages. The most challenging languages in terms of word prediction are Finnish and Turkish, with the model seemingly excelling at Portuguese, Mandarin, and English, to name a few. The lower scores observed for Finnish and Turkish can be explained by the small sample of training data in those languages, respectively 0.48\% and 1.57\% of the entire dataset, as well as by the complex morphology and productive compounding associated with these languages, as found by Buechel et al. \cite{emotion_lexicons_91_lang}. 

As for the short texts, compared in detail in Table \ref{tab:results_languages_s_texts}, the most challenging language was German. On this subject, we note that the German training sample contains the metaphorical instances of the COMETA dataset, which can explain the gap in the results for this language. Predicting valence in English also proved demanding. If analyzed in detail, the results are heavily influenced by the IEMOCAP dataset, which makes up for 46\% of the English short text corpus. IEMOCAP is a particular dataset, created through the video recording of actors performing scripts designed to contain select emotions~\cite{dataset_iemocap}. We used the transcriptions of the audio, which is annotated for valence and arousal in the dataset. Contrarily to all other datasets, these instances were annotated from videos, which can portray a large range of sentiments for the same textual script, depending on aspects such as posture and intonation of the actors. This implies that annotations range over a broader scope too, which likely affects the quality of the prediction results. 

Stemming from these last conclusions, we performed one more separate experiment. Considering the same training setting, we trained the model with a combined dataset not containing the two seemingly troublesome datasets, COMETA stories and IEMOCAP. Compared to previous results, the Pearson's $\rho$ for valence increased from $0.8095$ to $0.8423$, and arousal's correlation increased from $0.6974$ to $0.7107$. Performance gains were observed for all tested languages. In particular, valence and arousal correlation values for German short texts increased 13\% and 7\%, and most noticeably for English they increased 31\% and 11\%, respectively. This took the scores obtained for these two languages, which are well represented in the training instances, to levels akin to most other languages, and explained the previously noticed discrepancy in the evaluations.

\begin{table}[tb!]
 \centering
\caption{Evaluation results for individual languages on the short text datasets, using the XLM-RoBERTa-large model and considering the MSE loss function.}
\label{tab:results_languages_s_texts}
\scriptsize
\setlength{\tabcolsep}{9pt}
\begin{tabular}{lcccccc}
\hline
\footnotesize{\textbf{Language}} & \footnotesize{\boldmath$\mathrm{\rho_{V}}$} & \footnotesize{\boldmath$\mathrm{\rho_{A}}$} & \footnotesize{\boldmath$\mathrm{RMSE_{V}}$} & \footnotesize{\boldmath$\mathrm{RMSE_{A}}$} & \footnotesize{\boldmath$\mathrm{MAE_{V}}$} & \footnotesize{\boldmath$\mathrm{MAE_{A}}$} \\ \hline
English & 0.592 & 0.719 & 0.118 & 0.138 & 0.085 & 0.096 \\
Mandarin & 0.892 & 0.657 & 0.077 & 0.100 & 0.059 & 0.080 \\
German & 0.619 & 0.533 & 0.179 & 0.117 & 0.133 & 0.090 \\
Portuguese & 0.935 & 0.694 & 0.115 & 0.124 & 0.082 & 0.100 \\
Polish & 0.868 & 0.607 & 0.113 & 0.111 & 0.082 & 0.089 \\
\hline
\end{tabular}
\end{table}

\subsection{Results in Zero-Shot Settings}
With the previous results in mind, a question remained on whether our best model could generalize well to other languages in which it was not trained on. For that purpose, two other XLM-RoBERTa-large models were fine-tuned under the same training setup. Specifically, these models were trained with all the data from the merged dataset except for either the Polish or the Portuguese instances.
These instances were saved for subsequent zero-shot evaluations, separately focusing on each of these languages. 
This trial aimed to assert whether the proposed approach can generalize to a language not used for training. Polish and Portuguese were chosen for this purpose, as both these languages are represented in our dataset, simultaneously with word and short text instances. Despite being reasonably popular languages, they are not as extensively present as English, and thus they allow us to adequately simulate the scenario of testing the proposed model on a new language not seen during training, and also not seen extensively during the model pre-training stage (i.e., the DiltilBERT and XML-RoBERTa models, despite being multilingual, have seen much more English training data in comparison to other languages).

We can compare the results of these zero-shot experiments, presented in Table~\ref{tab:results_zero_shot}, with the results obtained for the Polish and Portuguese subsets of predictions presented previously in Table \ref{tab:results_datasets}.
When comparing correlation and error metrics, we found overall worse results. However, the difference is not significant, and the results are in fact higher than some of the observed results for other languages on which the model was fine-tuned on. 
The zero-shot performance for both languages shows promising prospects for the application of the proposed approach to different languages without available emotion~corpora.

\begin{table}[tb!]
 \centering
\caption{Zero-shot evaluation for Polish (PL) and Portuguese (PT) data, using the XLM-RoBERTa-large model and considering the MSE loss function.}
\label{tab:results_zero_shot}
\scriptsize
\setlength{\tabcolsep}{5.7pt}
\begin{tabular}{llcccccc}
\hline
\textbf{Training on} & \textbf{Predicting on}                                   & \boldmath$\mathrm{\rho_{V}}$ & \boldmath$\mathrm{\rho_{A}}$ & \boldmath$\mathrm{RMSE_{V}}$ & \boldmath$\mathrm{RMSE_{A}}$ & \boldmath$\mathrm{MAE_{V}}$ & \boldmath$\mathrm{MAE_{A}}$ \\ \hline
All &  & 0.839 & 0.648 & 0.101 & 0.103 & 0.072 & 0.080 \\
All excl. PL & \multirow{-2}{*}{Any PL input} & 0.818 & 0.618 & 0.111 & 0.135 & 0.080 & 0.108  \\
\hline 
All &  & 0.895 & 0.756 & 0.108 & 0.107 & 0.075 & 0.084    \\
All excl. PT & \multirow{-2}{*}{Any PT input} &  0.886 & 0.735 & 0.112 & 0.112 & 0.079 & 0.088 \\ \hline \hline
All &  &  0.833 & 0.631 & 0.100 & 0.102 & 0.071 & 0.079  \\
All excl. PL & \multirow{-2}{*}{PL words} &   0.814 & 0.647 & 0.111 & 0.135 & 0.079 & 0.108 \\
\hline 
All & & 0.893 & 0.779 & 0.106 & 0.103 & 0.074 & 0.081 \\
All excl. PT & \multirow{-2}{*}{PT words} &  0.906 & 0.777 & 0.102 & 0.107 & 0.071 & 0.084  \\ 
\hline \hline
All &  &  0.868 & 0.607 & 0.113 & 0.111 & 0.082 & 0.089  \\
All excl. PL & \multirow{-2}{*}{PL short texts} & 0.860 & 0.487 & 0.113 & 0.135 & 0.085 & 0.108   \\
\hline
All & & 0.935 & 0.694 & 0.115 & 0.124 & 0.082 & 0.100   \\
All excl. PT & \multirow{-2}{*}{PT short texts} & 0.923 & 0.627 & 0.155 & 0.135 & 0.121 & 0.109   \\ \hline
\end{tabular}
\end{table}

\section{Conclusions and Future Work}\label{section:conclusions}

This paper presented a bi-dimensional and multilingual model to predict real-valued emotion ratings from instances of text.
First, a multi-language emotion corpus of words and short texts was assembled. This goes in contrast to most previous studies, which focused solely on words or texts in a single language. The corpus, consisting of 128,987 instances, features annotations for the psycho-linguistic dimensions of Valence and Arousal (VA), spanning 13 different languages. Subsequently, DistilBERT and XLM-RoBERTa models were trained for VA prediction using the multilingual corpus.
The evaluation methodology used Pearson's $\rho$ and two error metrics to assess the results. Overall, the predicted ratings showed a high correlation with human ratings, and the results are in line with those of previous monolingual predictive approaches. Additionally, this research highlights the challenge of predicting arousal to the same degree of confidence of predicting valence from text.
In sum, the evaluation of our best model showed competitive results against previous approaches, having the advantage of generalization to different languages and different types of text. 

An interesting idea to explore in future work concerns applying uncertainty quantification\footnote{\url{https://mapie.readthedocs.io/en/latest/}} to the predicted ratings, for instance as explained by Angelopoulos and Bates~\cite{future_work_uncertainty_quantification}. Instead of predicting a single pair of values for the valence and arousal ratings, the aim would be to predict a high confidence interval of values in which valence and arousal are contained. Future work can also address the study of data augmentation methods (e.g., based on machine translation), in an attempt to further improve result quality and certainty.

Another interesting direction for future work concerns extending the work reported in this paper to consider multimodal emotion estimation. Instead of the models considered here, we can consider fine-tuning a large multilingual vision-and-language model\footnote{\url{https://huggingface.co/laion/CLIP-ViT-H-14-frozen-xlm-roberta-large-laion5B-s13B-b90k}} such as CLIP~\cite{carlsson2022cross}, combining the textual datasets together with affective image datasets like the International Affective Picture System (IAPS)~\cite{lang2005international}, the Geneva Affective PicturE Database (GAPED)~\cite{dan2011geneva}, the Nencki Affective Picture System (NAPS)~\cite{marchewka2014nencki}, the Open Affective Standardized Image Set (OASIS)~\cite{kurdi2017introducing}, or others~\cite{carretie2019emomadrid,kim2018building}.

\section*{Acknowledgements}

This research was supported by the European Union's H2020 research and innovation programme, under grant agreement No. 874850 (MOOD), as well as by the Portuguese Recovery and Resilience Plan (RRP) through project C645008882-00000055 (Responsible.AI), and by Funda\c{c}\~ao para a Ci\^encia e Tecnologia (FCT), through the INESC-ID multi-annual funding with reference UIDB/50021/2020, and through the projects with references DSAIPA/DS/0102/2019 (DEBAQI) and PTDC/CCI-CIF/32607/2017 (MIMU).

%
%
%
\bibliographystyle{splncs04}
\bibliography{references}

\end{document}